# Photo-realistic Facial Texture Transfer

Parneet Kaur   Hang Zhang   Kristin Dana

Department of Electrical and Computer Engineering, Rutgers University, New Brunswick, USA
`parneet@rutgers.edu, zhang.hang@rutgers.edu, kdana@ece.rutgers.edu`

## Abstract

Style transfer methods have achieved significant success in recent years with the use of convolutional neural networks. However, many of these methods concentrate on artistic style transfer with few constraints on the output image appearance. We address the challenging problem of transferring face texture from a style face image to a content face image in a photorealistic manner without changing the identity of the original content image. Our framework for face texture transfer (FaceTex) augments the prior work of MRF-CNN with a novel facial semantic regularization that incorporates a face prior regularization smoothly suppressing the changes around facial meso-structures (e.g eyes, nose and mouth) and a facial structure loss function which implicitly preserves the facial structure so that face texture can be transferred without changing the original identity. We demonstrate results on face images and compare our approach with recent state-of-the-art methods. Our results demonstrate superior texture transfer because of the ability to maintain the identity of the original face image.

## 1   Introduction

Recent work in texture synthesis and style transfer has achieved great success using convolutional neural networks [7, 8]. Despite the success of artistic style transfer, facial style transfer remains challenging due to the requirement of photo-realism and semantic consistency. Human vision is very sensitive to facial irregularities and even small distortions can make a face look unrealistic [16, 22]. In this work, we address the problem of photo-realistic facial style transfer, which transfers *facial texture* from a new style image while preserving most of the original *facial structure* and identity (Figure 1). Facial texture comprises skin texture details like wrinkles, pigmentation and pores, while facial structure consists of the meso-structures such as eyes, nose, mouth and face shape. Our approach has important implications in commercial applications and dermatology, such as visualizing the effects of age, sun exposure, or skin treatments (e.g. anti-aging, acne).

Style transfer of artistic work is typically approached by synthesizing a style texture based on the semantic content of the input image [4, 5, 12, 25]. Classic algorithms match the feature statistics of multi-scale representations [2, 9, 18]. Gatys et al. [6, 8] first adopted a pre-trained CNN [21] as a statistical feature representation to provide an explicit representation of image content and style. The output image is generated by solving an optimization problem which minimizes both content and style differences and iteratively passes the gradient directly to the image pixels. Recent work also explores real-time style transfer by training feed-forward networks while approximating the optimization process which outputs the style transferred images directly [11, 13, 24], and has been extended to multi-style [3, 26].

Despite the rapid growth of artistic style transfer work, photo-realistic facial style transfer remains challenging due to the need of preserving local semantic consistency while transferring skin texture. The Gram matrix is often used as a gold-standard style representation. Minimizing the difference of a global representation of Gram matrix does not sufficiently enforce local semantic consistency at meso-structures such as lower facial contour, eyes and mouth as shown in Figure 3 (last column).

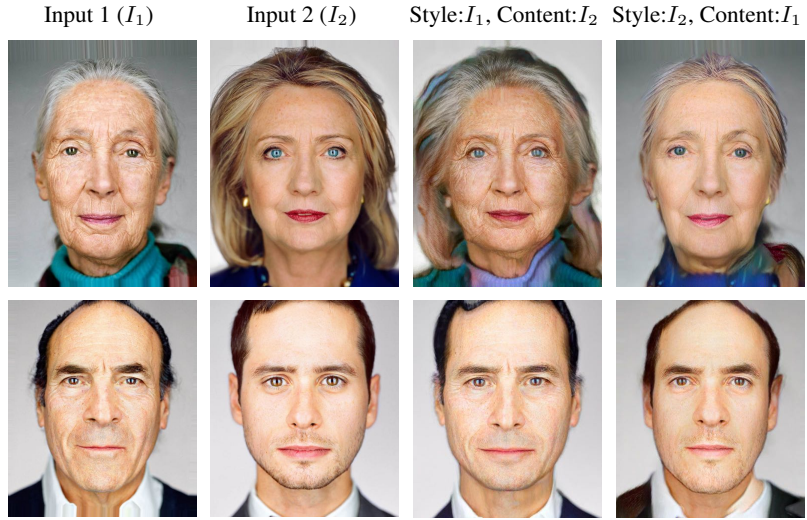

| Input 1 ($I_1$) | Input 2 ($I_2$) | Style:$I_1$, Content:$I_2$ | Style:$I_2$, Content:$I_1$ |

Figure 1: Identity-preserving Facial Texture Transfer (FaceTex). The textural details are transferred from style image to content image while preserving its identity. FaceTex outperforms existing methods perceptually as well as quantitatively. Column 3 uses input 1 as the style image and input 2 as the content. Column 4 uses input 1 as the content image and input 2 as the style image. Figure 3 shows more examples and comparison with existing methods. Input photos: Martin Scheoller/Art+Commerce.

A recent method [15] incorporates the Gram matrix with semantic segmentation and achieves high quality results for photo-realistic style transfer in scene images. This approach removes distortions in architectural scenes but is not designed for facial texture transfer and has no mechanism for retaining facial structure. Our approach is developed with the specific goal of maintaining the content face identity.

Markov random field (MRF) models have been used widely for representing image texture [27] by modeling the image statistic at a pixel or patch level and the dependence between neighbors. Classic texture synthesis methods using MRF [5][25] provide new texture instances using an MRF texture model. A recent work called MRF-CNN [13] leverages the local representation of MRF and the descriptive power of CNN for style transfer. However, this method also transfers meso-structures from the style image. For faces, this facial structure sourced from the style image leads to an undesirable change in facial identity during the texture transfer as in Figure 3 (column 3).

As the **first contribution** of this paper, we introduce *Facial Semantic Regularization* that consists of a *Facial Prior Regularization* and *Facial Structural Loss* for preserving identity during the texture transfer. Facial identity incorporates facial structure and shapes. We suppress the changes around the meso-structures by introducing the Facial Prior Regularization that smoothly slows down the updating. Additionally, we tackle the challenge of preserving facial shape by minimizing a Facial Structure Loss which we define as an identity loss from a pre-trained face recognition network that implicitly preserves the facial structure.

The **second contribution** of this paper is the development of an algorithm for *Identity Preserving Facial Texture Transfer* which we call *FaceTex* along with a complete benchmark of facial texture transfer with a novel metric for quantitative evaluation. Our approach augments the MRF-CNN framework with the Facial Semantic Regularization and faithfully transfers facial textures and preserves the facial identity. We provide a complete benchmark that evaluates style transfer algorithms on facial texture transfer task. Prior methods typically rely on perceptual evaluation of results, which makes it difficult to quantitatively compare them. We propose metrics that quantify the facial structure consistency as well as texture similarity. The experimental results show that the proposed FaceTex outperforms the existing approaches for identity-preserving texture transfer perceptually as well as quantitatively.

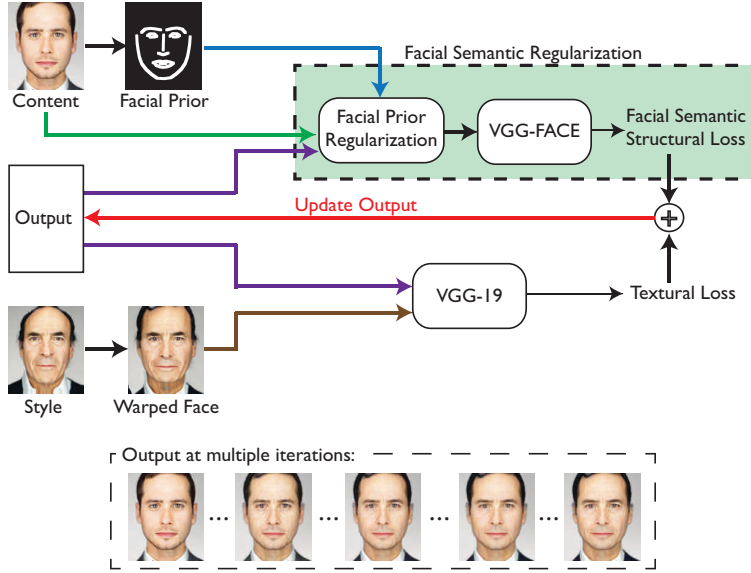

Figure 2: Overview of our method. Facial identity is preserved using Facial Semantic Regularization which regularizes the update of meso-structures using a facial prior and facial semantic structural loss. Texture loss regularizes the update of local textures from the style image. The output image is initialized with the content image and updated at each iteration by back-propagating the error gradients for the combined losses. Content/style photos: Martin Scheoller/Art+Commerce.

## 2 Methods

### 2.1 Texture Representation

We follow prior work of MRF-CNN [13] for texture representation and briefly describe it for completeness [13]. A pre-trained VGG-19 [21] is used as a descriptive representation of image statistics, and the feature-maps at layer $l$ for input image $x$ is denoted as $\Phi^l(x)$. For a given content image $x_c$ and a style image $x_s$, the facial texture is transferred from $x_s$ to the output/target image $x_t$ by minimizing the difference of local patches. Let $\Psi(\Phi^l(x))$ denote the set of the local patches on the featuremaps. For each patch $\Psi_i(\Phi^l(x_t))$, the difference with the most similar patch in the style image $\Psi_{NN(i)}(\Phi^l(x_s))$ (among $N_s$ patches) is minimized. The distance of the nearest neighbor is defined using normalized cross-correlation as

$$NN(i) = \underset{j=1,\ldots N_s}{\mathrm{argmin}} \frac{\Psi_i(\Phi^l(x_t))\Psi_j(\Phi^l(x_s))}{|\Psi_i(\Phi^l(x_t))| \cdot |\Psi_j(\Phi^l(x_s))|}. \quad (1)$$

The texture loss is the sum of the difference for all the $N_t$ patches in the generated image and is given by

$$\ell^l_{tex}(x_t, x_s) = \sum_{i=1}^{N_t} \|\Psi_i(\Phi^l(x_t)) - \Psi_{NN(i)}(\Phi^l(x_s))\|^2. \quad (2)$$

In contrast to the Gram Matrix that gives global impact to the image, MRF-CNN is good for preserving local textural structures. However, it also carries the semantic information from the style image, which violates the goal of preserving facial identity. For this, we augment the MRF-CNN framework with additional regularizations.

### 2.2 Facial Semantic Regularization

Facial identity consists of meso-structures including eyes, noise, eyebrow, lips and face contour. We tackle the problem of preserving facial identity by suppressing local changes around these meso-structures and minimizing the identity loss from face recognition network, which implicitly preserves the semantic facial structure.

**Facial Prior Regularization** Inspired by the dropout regularization [23] which randomly drops some units and blocks the gradient during the optimization, we build a facial prior regularization that smoothly slows down the updating around the meso-structures. For generating the facial prior mask, we follow the prior work [19] to generate 66 landmark points and draw contours for meso-structures. Then we build a landmark mask by applying a Gaussian blur to the facial contour and normalize the output between 0 and 1, which provides a smooth transition between meso-structures and rest of the face. For implementation, we build a CNN layer that performs an identity mapping during the forward pass of the optimization, and scales the gradient with an element-wise product with the face prior mask during back-propagation.

**Facial Semantic Structure Loss** Deep learning is well known for learning hierarchical representations directly from data. Instead of manually tackling preservation of facial structure, we minimize the perceptual difference of a face recognition network to force the output image to be recognized as the same person depicted in the input/content image. VGG-Face [17] is trained on millions of faces and has superior discriminative power for face recognition, which captures the facial meso-structures for identifying the person. Instead of minimizing the final classification error, we minimize the difference of mid-level feature-maps, because the mid-level features are already discriminative for preserving facial identity. Let $\delta^i(x)$ denote the feature-maps at a $i$-th layer of a pre-trained VGG-Face for input image $x$. The structure loss is the $L^2$-distance of the feature-maps and is given by

$$\ell_{face}(x_t, x_c) = \sum_{i=1}^{N_l} \frac{1}{C_i H_i W_i} \|\delta(x_t) - \delta(x_c)\|^2, \quad (3)$$

where $N_l$ is total number of layers for calculating structure loss, and $C_i$, $H_i$ and $W_i$ are the number of channels, height and width of the feature-map, respectively.

### 2.3 Identity Preserving Facial Texture Transfer

**Pre-processing** To maintain facial structural consistency and avoid artifacts, we warp the style image to the facial structure of the content image. First, 66 facial landmark points are generated for the content and style images using an existing facial landmark detection algorithm [19]. The style image is then morphed and aligned to the content image [1]. To further align the face contour we apply sift-flow, which uses dense SIFT feature correspondences for alignment while preserving spacial discontinuities [14].

**Loss functions** Reconstructing the image from the loss of highly abstracted pre-trained networks makes the image look unrealistic and noisy. We follow the prior work [11, 13, 26] which uses total variation regularization (TV loss) to encourage the smoothness of the output image $x$, which is given by the squared norm of the gradients:

$$\ell_{TV}(x) = \sum_{i,j} \left( (x_{i,j+1} - x_{i,j})^2 + (x_{i+1,j} - x_{i,j})^2 \right). \quad (4)$$

We use a weighted combination of texture loss, facial structure loss and TV loss to find the output estimate $\hat{x}_t$ as follows

$$\hat{x}_t = \underset{x_t}{\operatorname{argmin}} \sum_{l=1}^{L} \lambda_{tex}^l \ell_{tex}^l(x_t, x_s) + \lambda_{face} \ell_{face}(x_t, x_c) + \lambda_{TV} \ell_{TV}(x_t), \quad (5)$$

where $L$ is total number of layers for texture loss and $\lambda_{tex}^l$, $\lambda_{face}$ and $\lambda_{TV}$ are the balancing weights for texture loss, facial structure loss and TV loss. The optimization is performed by manipulating the the content image $x_c$ by iteratively updating the image pixels using an L-BFGS solver.

## 3 Experimental Results

### 3.1 Facial Style Transfer Benchmark

**Baseline Approaches.** We use the publicly available implementation of *Neural Style transfer* for comparison [8, 10]. Gatys et al. [8] generates an output image $x_t$ from content image $x_c$ and style image $x_s$ by jointly minimizing the content loss and the style loss iteratively. The content loss is given by the $L^2$-distance of the feature-maps at each convolution layers for the output and the content images. The style loss is the Frobenius norm of the Gram matrix difference of the feature-maps of output image and the style images at each layer. The weighted combination of the losses is minimized to obtain the output image as

$$\hat{x_t} = \underset{x_t}{\operatorname{argmin}} \lambda_c ||\Phi(x_t) - \Phi(x_c)||^2 + \lambda_s \sum_{l=1}^{L} ||\mathcal{G}^l(x_t) - \mathcal{G}^l(x_s)^l||_F^2 + \lambda_{TV}\ell_{TV}(x_t), \quad (6)$$

where $\Phi(x_t)$ and the $\Phi(x_c)$ are the feature-maps of output and style images, $\mathcal{G}^l(x_t)$ and $\mathcal{G}^l(x_s)$ are the Gram matrices of the feature-maps of output and style images at layer $l$; $L$ is the total number of layers; $\lambda_c$, $\lambda_s$ and $\lambda_{TV}$ are the weights for content loss, style loss and TV loss. In these experiments, we use $\lambda_c = 5$, $\lambda_s = 100$ and $\lambda_{TV} = 10^{-3}$. We use the L-BFGS solver for 1000 iterations. VGG-19 [21] pre-trained network is used for computing feature-maps. Layer relu4_2 is used for content loss while layers relu1_1,relu2_1,relu3_1,relu4_1 and relu5_1 are used for style loss.

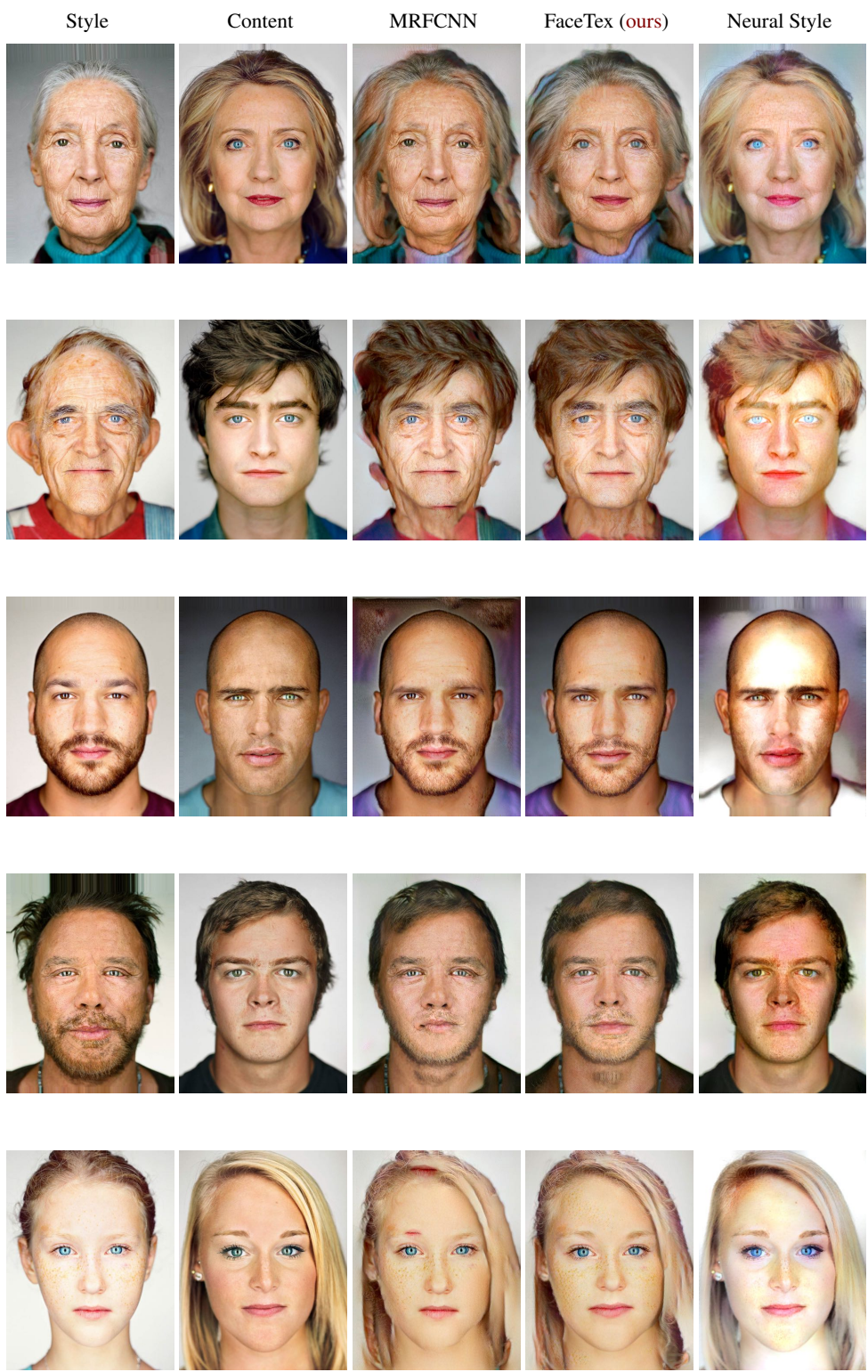

Figure 3: Our facial texture transfer on different content-style pairs. FaceTex (our approach, 4th column) preserves the identity of the content image and also transfers the textural details from the style image. Neural style (last column) preserves the identity but does not transfer the textural details. MRFCNN (3rd column) transfers the textural details but does not preserve the content image identity as well as FaceTex (compare 3rd and 4th column to content image in 2nd column). Content/style photos: Martin Scheoller/Art+Commerce.

We also compare our work with it MRF-CNN [13]. The output image is generated by minimizing the patch difference with the style image and preserving the high-level structure the same as in the content image. The loss function consists texture loss, content loss and TV losses:

$$\hat{x_t} = \underset{x_t}{\operatorname{argmin}} \lambda_c ||\Phi(x_t) - \Phi(x_c)||^2 + \lambda_s^l \sum_{l=1}^{L} \ell_{tex}^l(x_t, x_s) + \lambda_{TV}\ell_{TV}(x_t), \quad (7)$$

where $\ell_{tex}(x_t, x_s)$ is the texture loss as in Section 2.1, $\lambda_c$, $\lambda_s$ and $\lambda_{TV}$ are the wights for content loss, style loss and TV loss. Layers relu3_1 and relu4_1 of VGG-19 are used for texture loss and layer relu4_2 for content loss. Neural patches of size $3 \times 3$ are used to find the best matching patch. Three resolutions with 100 iterations each are used.

**Implementation Details.** We follow the work of MRF-CNN using layers relu3_1 and relu4_1 of VGG-19 [21] for texture loss. Layer relu4_2 of a pre-trained VGG-Face [17] is used for facial semantic structure loss. The facial prior mask is generated by connecting the landmark points using 40 pixel thickness line and applying a Gaussian blurring with the kernel size of 65 and standard divination of 30. In addition, the background mask provided in the dataset is also used. We incorporate facial prior regularization to block the changes of facial prior and background regions. We resize the content and style images to $1,000$ pixels along the long edge. The output image is initialized with the content image and the optimization is performed using the L-BFGS solver. We follow Li and Wand [13] using a multi-resolution process during the generation, the content and style images are scaled accordingly. We start with $\frac{1}{4}$ resolution and scale up by a factor of 2, and perform 200 iterations at each resolution. We use the same resolution for both baselines and our approach in this experiment.

**Metrics for Quantitative Evaluation.** We identify two metrics to quantitatively measure the facial structural inconsistency and texture similarity of the output image $x_t$ with the content image $x_c$ and the style image $x_s$.

*Landmark Error:* Using the methods described in section 2.3, we obtain $L = 66$ landmarks for each facial image. The output image has same facial structure if its landmark points remain the same as content image. The mean square error of the landmarks between the two images accounts for the facial structural inconsistency between them. Lower error indicates identity is preserved. The landmark error between two facial images is given by the $L^2$ distance of the pixel coordinates for the landmark points.

*Texture Correlation*: To measure the similarity between the output image and the input images, we can extract skin patches from the images and use the *normalized correlation coefficient*. Higher value of correlation coefficient indicates a better match of facial textures. Texture similarity of two patches $p$ and $q$ is given by:

$$S(p,q) = \frac{\sum_{ij}(p_{ij} - \bar{p})(q_{ij} - \bar{q})}{\sqrt{\sum_{i,j}(p_{ij} - \bar{p})^2 \sum_{i,j}(q_{i,j} - \bar{q})^2}}, \quad (8)$$

where $p_{ij}$ and $q_{ij}$ are the image values at pixel coordinates $(i,j)$ of the patches, $\bar{p}$ and $\bar{q}$ are the average pixel values of patches $p$ and $q$.

|  | **Neural Style** | **MRF-CNN** | **FaceTex (ours)** |
|---|---|---|---|
| $E(x_t, x_c)$ | 22.61 | 106.17 | 37.93 |
| $E(x_t, x_s)$ | 369.39 | 191.47 | 295.93 |
| $S(x_t, x_c)$ | 0.89 | 0.70 | 0.76 |
| $S(x_t, x_s)$ | 0.47 | 0.68 | 0.55 |

Table 1: Metrics for quantitative evaluation. The average metric values of the pairs in Figure 3 are reported here. For FaceTex, landmark error between output and content $E(x_t, x_c)$ is much lower than MRF-CNN indicating it is better at preserving identity. Texture similarity between output and style $S(x_t, x_s)$ is higher in FaceTex than Neural Style which shows that it is better in transferring texture.

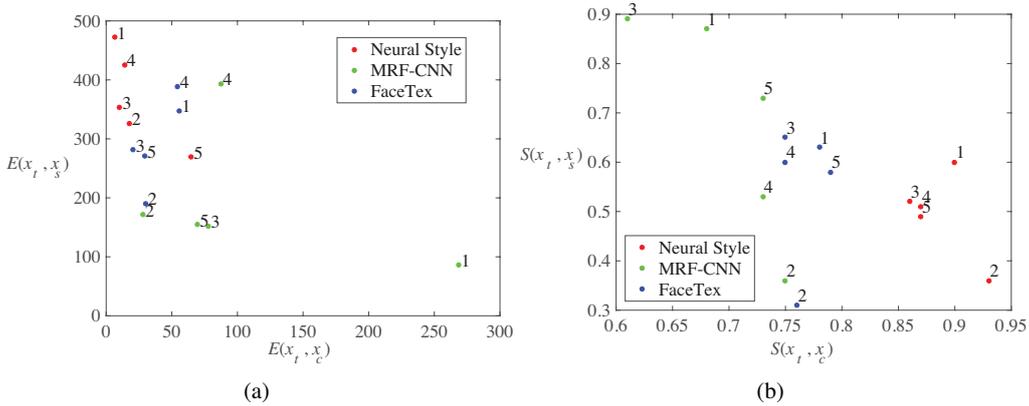

(a)  (b)

Figure 4: Quantitative evauation for each content-style pair. (a) Landmark Error. For FaceTex, $E(x_t, x_c)$ is small and much closer to Neural Style than MRF indicating that it preserves identity as in Neural Style approach. (b) Texture similarity. For FaceTex, $S(x_t, x_s)$ is high and closer to MRF-CNN, transferring tecture from style image.

### 3.2 Qualitative and Quantitative Comparison

We use the head portrait dataset provided by Shih *et al.* [20] for evaluation. Figure 3 shows the comparison of the output image generated using FaceTex with Neural Style and MRF-CNN. We provide additional comparison in the supplementary material. We observe that Neural Style preserves the facial structure and shape well but fails to transfer the texture, which demonstrates that the Gram Matrix transfers global styles well but fails to preserve the local finer texture and also makes the image unrealistic. MRF-CNN transfers local texture very well but it does not preserve the meso-structures which leads to more significant change the observed facial identity. Our proposed FaceTex approach generates photo-realistic images and outperforms all the baseline approaches in transferring facial texture as well as preserving the facial identity.

The quantitative comparison matches the conclusion of qualitative observation, and the results of landmark and texture metrics are listed in Table 1. The average values of different metrics are reported for the content-style pairs in Figure 3. $E(x_t, x_c)$ and $E(x_t, x_s)$ are the landmark errors of output image with content and style images, respectively. $E(x_t, x_c)$ is very low for Neural Style (22.61) but high for MRF-CNN (106.17) indicating that identity is preserved by the Neural Style approach. For FaceTex, $E(x_t, x_c)$ is much lower (37.93) than MRF-CNN and preserves the identity. In contrast, higher value of $E(x_t, x_s)$ indicates that facial structural similarity is not maintained with the style image as expected. $S(x_t, x_c)$ and $S(x_t, x_s)$ are the texture similarities of output image with content and style images, respectively. For each content-style pair, we extract three patches and report their average normalized correlation coefficient as the texture similarity of the image. These patches are $100 \times 100$ and in forehead and both cheek regions, localized by face structure landmarks. For texture transfer, a large value of $S(x_t, x_s)$ indicates that texture is successfully transferred to output image from the style image. $S(x_t, x_s)$ is highest for MRF-CNN (0.68) and lowest for Neural Style (0.47), whereas for FaceTex similarity lies between MRF-CNN and Neural Style (0.55).

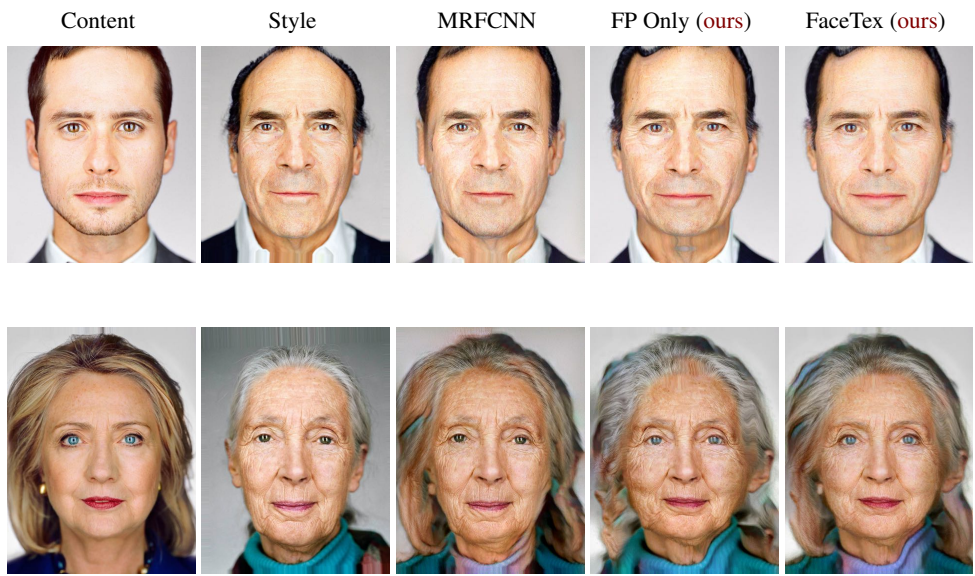

Figure 5: The effects of facial prior (FP) regularization and facial semantic structure loss. Using FP regularization (column 4) preserves better meso-structure of the faces comparing to MRF-CNN (column 3). Facial semantic loss effectively preserve the facial structure for identity preserving as shown in the last column. Content/style photos: Martin Scheoller/Art+Commerce.

Figure 4(a) and (b) shows the landmark errors and texture similarity for each of the five content-style pairs in Figure 3. Both the error and the similarity measures for FaceTex (blue dots) lie between Neural Style (red dots) and MRF-CNN (green dots), and generally much closer to MRF-CNN.

**Ablation Experiments.** Figure 5 exemplifies the necessity of augmenting the existing methods with multiple regularizations. If only the facial prior regularization is used, the generated output face still looses identity and has artifacts. Adding the facial semantic structure further preserves the identity and suppresses some artifacts.

**Limitations.** Our method achieves superior performance in identity preserving facial texture transfer and generates photo-realistic images, but still has its limitations. First, our approach is an optimization-based approach, which takes several minutes generating a new image, which limits the applications in real-time. This could be potentially addressed in the future work by combining a feed-forward network and a face alignment network that run in real-time. Second, the texture modeling using MRF-CNN requires high semantic similarity between two input images, which may lead to some unappealing artifacts for mismatches.

## 4 Conclusion

We have presented the method FaceTex for photo-realistic facial style transfer. By augmenting prior work of MRF-CNN with a novel regularization consisting of a facial prior regularization and the facial semantic structure loss, we can transfer texture realistically while retaining semantic structure so that the identity of the individual remains recognizable. Our results show substantial improvement over the state-of-the-art both in the quality of the texture transfer and the preservation of the original face structure. Quantitative metrics of texture transfer and face structure are also improved using this approach.

## Acknowledgments

We would like to thank Martin Scheoller/Art+Commerce for giving us permission to use their photographs in this paper.